\documentclass{article}
\usepackage{ijcai25}

\usepackage{times}
\usepackage{soul}
\usepackage{url}
\usepackage[hidelinks]{hyperref}
\usepackage[utf8]{inputenc}
\usepackage[small]{caption}
\usepackage{graphicx}
\usepackage{amsmath}
\usepackage{amsthm}
\usepackage{booktabs}
\usepackage{algorithm}
\usepackage{algorithmic}
\usepackage[switch]{lineno}

\usepackage{subfigure}
\usepackage{multirow}
\usepackage{threeparttable}
\usepackage{enumitem}
\usepackage{caption}
\usepackage{xcolor}
\usepackage{float} 
\usepackage{booktabs}
\usepackage{arydshln}
\usepackage{algorithm}
\usepackage{algorithmic}
\usepackage{cleveref}
\usepackage{amssymb}
\crefname{table}{Table}{Tables}

\title{
Unveiling the Power of Noise Priors: Enhancing Diffusion Models \\ for  Mobile Traffic Prediction

}

\author{
Zhi Sheng$^{1}$\thanks{\parbox[t]{0.9\textwidth}{Equal contribution.\\ The full version is at \url{https://arxiv.org/pdf/2501.13794}.}}
\and
Daisy Yuan$^{1}$\footnotemark[1]\and
Jingtao Ding$^{1}$\and Qi Yan$^{2}$\and Xi Zheng$^{2}$\and Yue Sun$^{2}$\And
Yong Li$^{1}$\\
\affiliations
$^{1}$Department of Electronic Engineering, Tsinghua University\\
$^2$Huawei Technologies Co., Ltd\\
\emails
liyong07@tsinghua.edu.cn
}

\begin{document}
\maketitle

\begin{abstract}

Accurate prediction of mobile traffic, \textit{i.e.,} network traffic from cellular base stations, is crucial for optimizing network performance and supporting urban development. However, the non-stationary nature of mobile traffic, driven by human activity and environmental changes, leads to both regular patterns and abrupt variations. 
Diffusion models excel in capturing such complex temporal dynamics due to their ability to capture the inherent uncertainties. 
Most existing approaches prioritize designing novel denoising networks but often neglect the critical role of noise itself, potentially leading to sub-optimal performance.
In this paper, we introduce a novel perspective by emphasizing the role of noise in the denoising process. 
Our analysis reveals that noise fundamentally shapes mobile traffic predictions, exhibiting distinct and consistent patterns.
We propose NPDiff, a framework that decomposes noise into \textit{prior} and \textit{residual} components, with the \textit{prior} derived from data dynamics, enhancing the model's ability to capture both regular and abrupt variations.
NPDiff can seamlessly integrate with various diffusion-based prediction models, delivering predictions that are effective, efficient, and robust. Extensive experiments demonstrate that it achieves superior performance with an improvement over 30\%, offering a new perspective on leveraging diffusion models in this domain. We provide code and data at \textcolor{blue}{\url{https://github.com/tsinghua-fib-lab/NPDiff}}.
\end{abstract}

\section{Introduction}
\begin{figure}[t]
    \centering
    \includegraphics[width=0.95\linewidth]{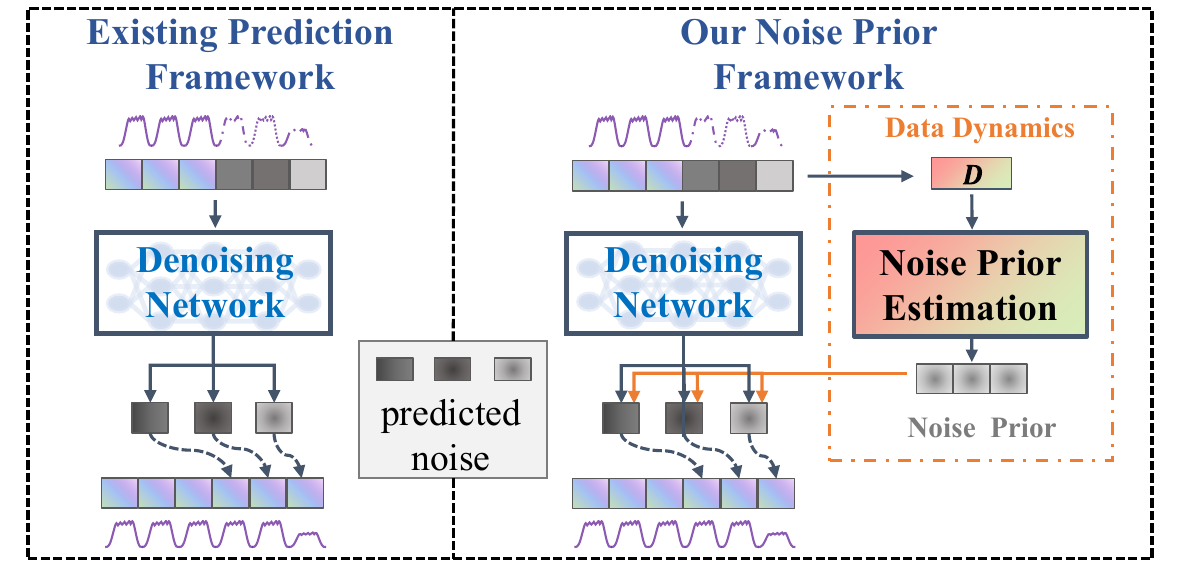}
    \caption{Comparison of existing diffusion-based prediction framework with our noise prior framework.} 
    \label{fig:intro}
    \vspace{-1mm}
\end{figure}

Communication networks serve as the essential infrastructure of smart cities, facilitating data exchange and supporting various applications. With the rapid growth of network traffic driven by advanced communication technologies and the proliferation of smart devices, efficiently managing mobile traffic resources has become more challenging~\cite{sarker2021mobile}. Accurate traffic prediction is crucial for optimizing resource allocation, allowing operators to adjust bandwidth and base station power in response to fluctuating demand, thereby preventing congestion and reducing energy consumption~\cite{yu2022network}.

Mobile traffic prediction involves forecasting the dynamics of time series and spatio-temporal data. Traditional deep learning methods—such as 
CNNs~\cite{li2018diffusion,zhang2017deep}, RNNs~\cite{wang2017predrnn,lin2020self}, 
GCNs~\cite{zhao2019t,bai2020adaptive}, and transformers~\cite{yuan2024unist,yuan2024foundation}
—have achieved notable success in this area, often employing a deterministic approach that directly maps inputs to outputs. Recently, diffusion models~\cite{ho2020denoising,nichol2021improved,yuan2024urbandit} have emerged as a promising alternative. Unlike traditional models~\cite{yi2024get,zhou2024coms2t}, they offer a probabilistic framework that captures the intricate distributions inherent in temporal data, effectively handling uncertainties and complexities. For time series and spatio-temporal data, most diffusion model-based solutions~\cite{tashiro2021csdi,shen2023non,yuan2023spatio} naively adopt the noise modeling in the image domain and design specialized denoising networks. This noise is typically modeled as independent and identically distributed (i.i.d.) Gaussian noise, which guides the denoising process step-by-step.

However, mobile traffic data exhibits unique dynamics that differ significantly from those in images. Originating from human movements and activities, it naturally reflects human rhythms, with temporal dynamics often showing periodicity and abrupt changes~\cite{xu2016understanding,xu2016big}.
Existing diffusion model-based solutions mainly focus on designing innovative denoising networks or conditioning mechanisms to incorporate various features~\cite{tashiro2021csdi,shen2023non}. While conditioning features for the denoising network can provide some context for predictions, the noise learned in the denoising process step by step plays a more critical role~\cite{ge2023preserve,zhang2024trip}—it directly shapes the model’s generative process, significantly influencing the quality and accuracy of predictions. Despite its importance, the impact of noise in the denoising process for temporal data remains largely unexplored.

Therefore, we are motivated to address an essential research question: can the noise estimation process be re-designed to effectively incorporate the inherent characteristics of flow data?
However, it is not a trivial task.  First, the assumption that the noise to be predicted is Gaussian provides essential properties that ensure the stability and convergence of diffusion models. Manipulating this noise requires preserving these characteristics to maintain the model’s reliability. Second, mobile traffic data is highly non-stationary, exhibiting fluctuations driven by diverse factors like human activity patterns and environmental changes. As a result, the model must capture regular, predictable patterns while remaining sensitive to sudden changes.

To address these challenges, we introduce NPDiff, a novel \underline{\textbf{N}}oise \underline{\textbf{P}}rior framework for \underline{\textbf{Diff}}usion models in mobile traffic prediction, which leverages the patterns of data dynamics as priors to estimate noise, as illustrated in Figure~\ref{fig:intro}. The key idea is to decompose the noise into two components: (1) \textbf{\textit{noise prior}} captures the intrinsic dynamics of mobile traffic data, including periodic patterns and local variations, which provides a basic reference for the noise estimation in each step; (2) \textbf{\textit{noise residual}} accounts for the additional variations that the noise prior cannot fully represent, enabling the model to adapt to more complex and unpredictable aspects of the data.  
Based on theoretical derivations, we can successfully reconstruct mobile traffic data through the backward diffusion process using the noise prior and the noisy data.
As a general solution, NPDiff can be seamlessly integrated into existing diffusion-based models, boosting their prediction performance and adaptability to the unique characteristics of mobile traffic.
Our contributions are summarized as follows:

\begin{itemize}[leftmargin=*]
    \item To our best knowledge, we are the first to highlight noise's role in diffusion models for mobile traffic prediction, offering new insights to apply diffusion models in this domain.
    \item We propose NPDiff, a general framework for diffusion-based mobile traffic prediction. By leveraging the intrinsic dynamics of crowd data as noise priors, it can be seamlessly integrated into advanced diffusion models.
    \item Extensive experiments across various prediction tasks and different denoising networks demonstrate that NPDiff boosts prediction accuracy by over 30\%. Moreover, NPDiff improves training efficiency and robustness while reducing prediction uncertainty, positioning it as an efficient and versatile enhancement to current methodologies.
\end{itemize}

\section{Related Work}
\subsection{Diffusion Models for Spatiotemporal Data}
Diffusion models~\cite{nichol2021improved,ho2020denoising} have become popular for time series~\cite{tashiro2021csdi,shen2023non} and spatio-temporal data~\cite{wen2023diffstg,yuan2024spatiotemporal} due to their strong generative capabilities. 
These models utilize additional data, such as historical observations~\cite{shen2023non} or knowledge graphs~\cite{zhou2023towards}, to guide the denoising process. 
Currently, most diffusion-based approaches~\cite{tashiro2021csdi,shen2023non,wen2023diffstg,yuan2024spatiotemporal,yuan2023spatio} focus on the conditioning mechanism, with limited attention to optimizing the noise component. We address this gap by investigating the impact of noise and designing a noise prior framework, leading to more accurate predictions.

\subsection{Noise Manipulation in Video Models}
Diffusion models~\cite{ho2022video,blattmann2023stable} have made significant progress in video generation, but generating videos with consistency remains a core challenge~\cite{xing2023survey,chang2024warped}. Recently, many studies~\cite{chang2024warped,qiu2023freenoise,ge2023preserve,zhang2024trip} have enhanced the video coherence by manipulating noise during the diffusion process. Similar noise is sampled for each video frame in the early stage of diffusion~\cite{qiu2023freenoise,ge2023preserve}. New forms of noise representation are also introduced~\cite{chang2024warped}. Moreover, some researchers use historical images as references to provide static noise prior for the diffusion process~\cite{zhang2024trip}. Compared with video data that requires temporal consistency, mobile data are much more dynamic and complex. How to design effective noise priors for such data remains unexplored yet is a promising direction.

\section{Preliminaries}
\subsection{Problem Formulation}
Mobile traffic prediction uses historical data to predict future traffic flows. The format of mobile traffic data is characterized as a three-dimensional tensor  $X\in\mathbb{R}^{T \times K \times C}$, where $T$ is the temporal length, $K$ is the number of spatial locations, and $C$ denotes traffic flow features. Specifically, our task is to learn a predictive model $\mathcal{F}$, which, given historical context data $X^{co}=X^{t-H+1:t}$ of length $H$, predicts the traffic flow $X^{ta}=X^{t+1:t+M}$ for the next $M$ steps, which can be represented as $X^{ta}=\mathcal{F}(X^{co})$.

\subsection{Diffusion-based Mobile Traffic Prediction}

For mobile traffic prediction, the diffusion model predicts target values conditioned on historical data $x_0^{co}$. 
During the forward process, noise is progressively added to the target data with the noise schedule \(\{\beta_n\}_{t=1}^{N}\). Given \(\alpha_n=1-\beta_n\) and \(\bar{\alpha}_n=\prod_{i=1}^{n}\alpha_i\), the noised target at any diffusion step can be calculated using the formula below: 
\begin{equation}
\label{eq:one-setp-forward}
    x_n^{ta} = \sqrt{\bar{\alpha}_n} x_0^{ta} + \sqrt{1 - \bar{\alpha}_n} \epsilon, \quad \epsilon \sim \mathcal{N}(0, I).
\end{equation}
Reversely, the denoising process is formulated  as follows:

\begin{equation}
\label{eq:inference}
\begin{aligned}
    &p_{\theta}(x_{0:N}^{ta}) := p(x_N^{ta}) \prod_{n=1}^{N} p_{\theta}(x_{n-1}^{ta} | x_n^{ta}, x_0^{co}), \\
    &p_{\theta}(x_{n-1}^{ta} | x_n^{ta}) := \mathcal{N}(x_{n-1}^{ta}; \mu_{\theta}(x_n^{ta}, n| x_0^{co}), \Sigma_{\theta}(x_n^{ta}, n)),\\
    &\mu_{\theta}(x_n^{ta}, n| x_0^{co})=\frac{1}{\sqrt{\bar{\alpha}_n}} \left( x_n^{ta} - \frac{\beta_n}{\sqrt{1 - \bar{\alpha}_n}} \epsilon_{\theta}(x_n^{ta}, n| x_0^{co}) \right),\\
     &\Sigma_{\theta}(x_n^{ta}, n)=\frac{1 - \bar{\alpha}_{n-1}}{1 - \bar{\alpha}_n} \beta_n.
\end{aligned}
\end{equation}
The model is trained to estimate the noise $\epsilon$. The model's parameters $\theta$ are optimized by the following loss function:
\begin{equation}
\label{eq:loss1}
   \min_{\theta}\mathcal{L}(\theta)  = \min_{\theta} \mathbb{E}_{n, x_0, \epsilon} \left[ \left\| \epsilon - \epsilon_{\theta}(x_n^{ta}, n| x_0^{co}) \right\|_2^2 \right].
\end{equation}

\begin{figure}[t]
    \includegraphics[width=1\linewidth]{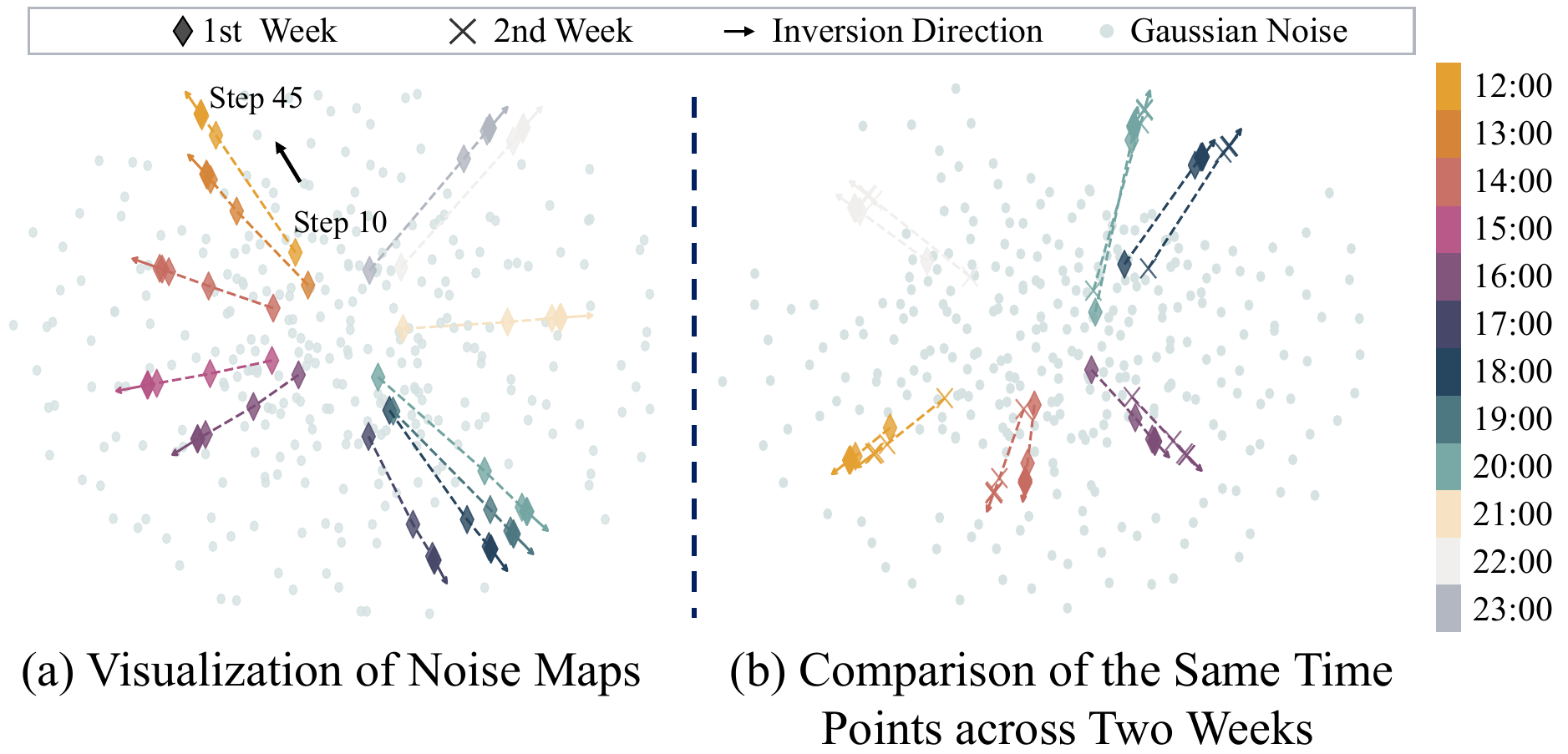}
    \caption{Noise visualization of the MobileNJ dataset.  (a) Noise in 12 consecutive time points during the denoising process, encompassing five distinct diffusion steps from step 10 to step 45.  (b) Noise comparison of the same timestamps across two consecutive weeks.}
    \label{fig:analysis}
    \vspace{-4mm}
\end{figure}

\section{Observations and Analysis} \label{section: analy}

To better demonstrate the role of noise in the denoising process, we visualize the estimated noise for different diffusion steps and timestamps using MobileNJ dataset. Specifically, we obtain the noise maps through the Denoising Diffusion Implicit Models (DDIM) inversion process and employ t-distributed Stochastic Neighbor Embedding (t-SNE) to visualize them.
The results are shown in Figure~\ref{fig:analysis}, the diamond squares denote the noise estimated by the model during this process while the other small dots in the background represent Gaussian noise, serving as a reference noise.

Figure~\ref{fig:analysis}(a) illustrates that, for a specific time point, the noise across different diffusion steps exhibits high continuity, approximately following a linear trend. This suggests that noise at every diffusion step is constrained by certain inherent conditions. Additionally, the noise shows a counterclockwise rotation trend, aligning with the temporal dimension. In Figure~\ref{fig:analysis}(b), we further observe that the noise from the same time point one week later highly overlaps with those from the current time point, revealing clear periodic characteristics. This periodicity is highly consistent with the inherent dynamics of mobile traffic data, indicating a strong connection between temporal dynamics and noise within diffusion process.

These findings underscore the critical role of noise in the denoising process, as the noise exhibits clear and consistent patterns. Consequently, integrating the intrinsic dynamics of temporal data into the modeling of noise is well-motivated and promising for enhancing the model's effectiveness.

\begin{figure*}[t]
    \centering
    \includegraphics[width=0.87\linewidth]{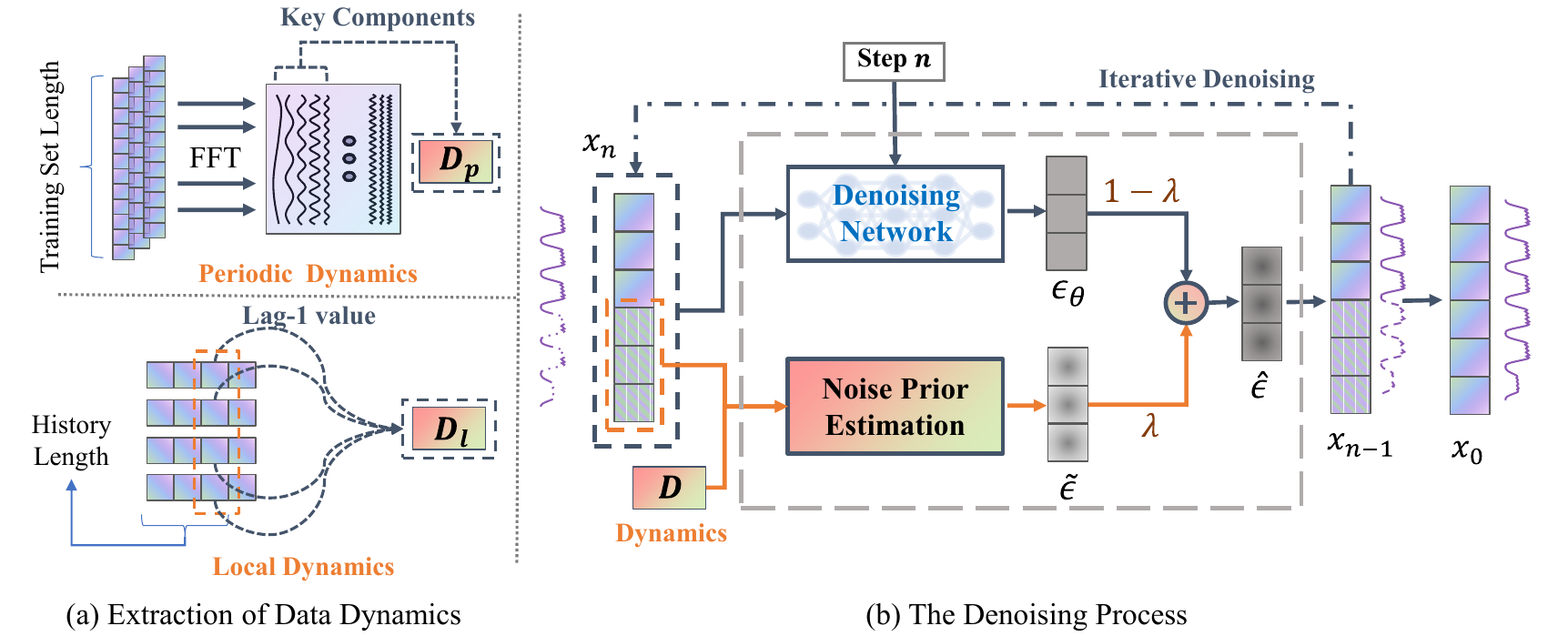}
    \vspace{-1mm}
    \caption{Overview of NPDiff. (a) shows the process of extracting data dynamics, and (b) shows the denoising process with noise priors.} 
    \label{fig:model}
\end{figure*}

\section{Method}
\subsection{Framework Overview}
We introduce the NPDiff framework, which incorporates the mobile traffic dynamics as noise prior to the diffusion process. Figure~\ref{fig:model} illustrates an overview of our approach. 
Figure~\ref{fig:model}(a) shows the extraction of two key data dynamics: periodic dynamics and local dynamics, while Figure~\ref{fig:model}(b) illustrates the denoising process with noise priors. 
The noise is estimated through two pathways: (1) directly calculating the noise prior based on data dynamics; and (2) predicting the residual noise using the denoising network.

\subsection{Dynamics of Mobile Traffic}
By analyzing the dynamics of mobile traffic data, we select two representative dynamics to calculate noise priors, which we define as periodic dynamics and local dynamics, respectively. Periodic dynamics capture the regular patterns, while local dynamics reflect short-term fluctuations.
We calculated the cosine similarities between these dynamics and actual values. As shown in Table~\ref{tbl:cos_similarity}, they both achieve high cosine similarities of around 0.8 across all datasets.

\noindent\textbf{Periodic Dynamics.} To capture periodic dynamics, we leverage Fast Fourier Transform (FFT), which is widely used in time series analysis. 
Specifically, we select several key components obtained by using FFT and convert them back to the time domain. This process can be formulated as follows:

\begin{equation}
    \begin{aligned}
    & A_{(k)} = \left| \text{FFT}(x)_k \right|, \quad \Phi_{(k)} = \mathbf{\phi} \left( \text{FFT}(x)_k \right), \\
    & \kappa^{(1)}, \cdots, \kappa^{(K)} = \underset{k \in \left\{ 1, \cdots, \left\lfloor \frac{L}{2} \right\rfloor + 1 \right\}}{\text{arg-TopK}} \left\{ A_{(k)} \right\}, \\
    & S[i] = \sum_{k=1}^{N_K} A_{\kappa^{(k)}} \Big[ \cos \left( 2\pi f_{\kappa^{(k)}} i + \Phi_{\kappa^{(k)}} \right) \\
    & \qquad \qquad + \cos \left( 2\pi \bar{f}_{\kappa^{(k)}} i + \bar{\Phi}_{\kappa^{(k)}} \right) \Big],
    \end{aligned}
\end{equation}

\noindent where $A_{(k)}$ is the amplitude of the 
$k-$th frequency component obtained by applying FFT to the training data $x$, and $\Phi_{(k)}$ represents the phase of the $k-$th frequency component. $L$ represents the length of the training set. $\kappa^{(1)}, \cdots, \kappa^{(K)}$ are the the top 
$K$ amplitudes selected by \text{arg-TopK}. $N_K$ represents the number of selected components, which is set to 5 or determined by the number of components with amplitudes exceeding the average amplitude in our paper. $f_{\kappa^{(k)}}$ refers to the frequency corresponding to the $k$-th component, and $\bar{f}_{\kappa^{(k)}}$, $\bar{\Phi}_{{\kappa^(k)}}$ represent the corresponding conjugates. We then compute the average values of the time-domain signal at corresponding time points across different cycles within a specified period length \(P\) to extract the final dynamics.  It can be formulated as follows: 
\begin{equation}
    D_p[t] = \frac{1}{N_{P}} \sum_{k=0}^{N_P - 1} S[t + kP],
\end{equation}
\noindent where \(N_{P}\) denotes the total number of complete periods in the training data, \(t+kP\) represents the sampled value at time \(t\) within each period \(P\). we extend periodic dynamics $D_p[t]$ within period \(P\) to the entire dataset to obtain the dynamics corresponding to each time point. 
The period \(P\) is defined as one week in most datasets, while \(P\) is set to one day for MobileBJ dataset due to the limited data duration. If the dataset is sufficiently large, \(P\) can be chosen to represent longer periods, such as a month or a year, which correspond to the behavioral patterns of human activity.

\begin{table}[t]
\centering
\vspace{-2mm}
\begin{threeparttable}
\resizebox{\columnwidth}{!}{
\begin{tabular}{lcccc}
\toprule
& \textbf{MobileBJ} & \textbf{MobileNJ} & \textbf{MobileSH14} & \textbf{MobileSH16} \\
\midrule
\textbf{Periodic Dynamics} & 0.889 & 0.876 & 0.778 & 0.763 \\
\textbf{Local Dynamics} & 0.923 & 0.905 & 0.826 & 0.825 \\
\bottomrule
\end{tabular}}
\end{threeparttable}
\caption{Cosine similarities between extracted dynamics and targets.}
\label{tbl:cos_similarity}
\vspace{-2mm}
\end{table}



\noindent\textbf{Local Dynamics.} Periodic dynamics reflect the periodic patterns, but they may not be well-suited for capturing short-term variations. Mobile traffic data usually depend on consecutive time steps, the current values are strongly correlated with the nearby last step. Therefore, we incorporate local dynamics.
Specifically, we utilize the lag-1 values to depict local variations. Mathematically, this can be described as: 
\begin{equation}
    D_l[t]=x[t-1].
\end{equation}

\subsection{Derivation and Fusion of Noise Priors}
As analyzed in Section~\ref{section: analy}, there is a strong correlation between noise and data dynamics. Therefore, our objective is to enhance the model's performance by leveraging these correlations. In this section, we describe the method for calculating noise priors and explain how they are integrated into the denoising process. 
For an intuitive representation of the diffusion process and the use of noise priors, please refer to Appendix~Figure~\ref{fig:revision_sup}.

\subsubsection{Derivations of Noise Priors}
Given the extracted periodic and local dynamics, a basic assumption is that they are closely aligned with the corresponding target values. Based on this assumption~\cite{zhang2024trip}, we can reformulate the mobile traffic data as follows:

\begin{equation}
\label{eq:prior-target}
    x_0^{ta} = D+\Delta x,
\end{equation}
\noindent where \(\Delta x\) represents a small error term that quantifies the difference between the target value and the extracted dynamics. 
We then rearrange Eq.~(\ref{eq:one-setp-forward}) to derive the formula below:
\begin{equation}
\label{eq:epsilon}
    \epsilon=\frac{x_n^{ta}-\sqrt{\bar{\alpha}_n}x_0^{ta}}{\sqrt{1-\bar{\alpha}_n}}.
\end{equation}
This noise $\epsilon$ serves as the target noise we need to estimate during the training process. Next, we substitute Eq.~(\ref{eq:prior-target}) into Eq.~(\ref{eq:epsilon}), and obtain:
\begin{equation}
\begin{aligned}
\label{eq:reference}
    \epsilon&=\frac{x_n^{ta}-\sqrt{\bar{\alpha}_n}(D+\Delta x)}{\sqrt{1-\bar{\alpha}_n}}\\
    &=\frac{x_n^{ta}-\sqrt{\bar{\alpha}_n}D}{\sqrt{1-\bar{\alpha}_n}}-\frac{\sqrt{\bar{\alpha}_n}}{\sqrt{1-\bar{\alpha}_n}}\Delta x.\\
\end{aligned} 
\end{equation}
Thus, the noise prior can be defined as follows: 
\begin{equation}
\label{eq:noise prior}
\widetilde{\epsilon} = \begin{cases}
   \frac{x_n^{ta}-\sqrt{\bar{\alpha}_n}D_l}{\sqrt{1-\bar{\alpha}_n}},  &\text{for one-step prediction} \\
   \frac{x_n^{ta}-\sqrt{\bar{\alpha}_n}D_p}{\sqrt{1-\bar{\alpha}_n}},  &\text{for multi-step prediction}
\end{cases}.
\end{equation}

According to Eq~(\ref{eq:noise prior}), the noise prior can be directly computed at any diffusion step based on data dynamics. And the target noise can be written as:
\begin{equation}
\begin{aligned}
    \epsilon&=\widetilde{\epsilon}-\frac{\sqrt{\bar{\alpha}_n}}{\sqrt{1-\bar{\alpha}_n}}\Delta x\\
    &=\widetilde{\epsilon} + \Delta\epsilon,
\end{aligned}
\end{equation}
where $\Delta\epsilon$ represents a small term, we term it `residual noise'.

\begin{table*}[t!]
\centering
\setlength\tabcolsep{4mm}


\begin{threeparttable}
\vspace{-2mm}
\resizebox{1.7\columnwidth}{!}{
\begin{tabular}{ccccccccc}
\toprule
\multirow{2}{*}{\textbf{Model}}
& \multicolumn{2}{c}{\textbf{MobileBJ}} & \multicolumn{2}{c}{\textbf{MobileNJ}} & \multicolumn{2}{c}{\textbf{MobileSH14}} & \multicolumn{2}{c}{\textbf{MobileSH16}}  \\
\cmidrule(lr){2-3} \cmidrule(lr){4-5} \cmidrule(lr){6-7} \cmidrule(lr){8-9} 
 &\textbf{MAE} & \textbf{RMSE} & \textbf{MAE} & \textbf{RMSE} & \textbf{MAE} & \textbf{RMSE} & \textbf{MAE} & \textbf{RMSE} \\
\midrule
HA & 0.232 & 0.343 & 0.454 & 0.881 & 0.100 & 0.165 & 13.44 & 38.92 \\
ARIMA & 0.236 & 0.404 & 0.327 & 0.692 & 0.112 & 0.197 & 9.15 & 26.70 \\
\midrule
PatchTST & 0.189 & 0.291 & 0.213 & 0.447  & 0.062 & 0.108 & 10.69 & 28.17 \\
iTransformer & 0.154 & 0.249 & 0.205 & 0.436 & 0.045 & 0.072  & 10.19 & 25.91 \\
Time-LLM & 0.115 & 0.195 & 0.200 & 0.394 & 0.060 & 0.102 & 10.57 & 28.19 \\
\midrule
STResNet & 0.546 & 0.751 & 0.439 & 0.624 & 0.102 & 0.138 & 45.63 & 59.82 \\
ATFM & 0.141 & 0.200 & 0.309 & 0.492 & 0.055 & 0.078 &24.95 & 46.92 \\
STNorm & 0.132 & 0.198 & 0.194 & 0.316 & 0.042 & 0.065 & 11.88 & 28.46 \\
STGSP & 0.157 & 0.229 & 0.214 & 0.323 & \underline{0.040} & \underline{0.057} & 17.54 & 38.77 \\

TAU & 0.135 & 0.196 & 0.268 & 0.389 & 0.044 & 0.063 & 15.22 & 26.04 \\
PromptST & \underline{0.099} & \textbf{0.171} & \underline{0.157} & \underline{0.303} & 0.043 & 0.069 & \underline{9.30} & \underline{23.01} \\
MAU & 0.166 & 0.256 & 0.387 & 0.662 & 0.081 & 0.125 & 21.38 & 45.04 \\
MIM & 0.214 & 0.298 & 0.270 & 0.447 & 0.079 & 0.126 & 22.49 & 47.29 \\
\midrule
CSDI & 0.596 & 3.178 & 0.173 & 0.316 & 0.045 & 0.072 & 10.35 & 41.78 \\
\textbf{CSDI+Prior} & \textbf{0.094} & \underline{0.173} & \textbf{0.096} & \textbf{0.158} & \textbf{0.037} & \textbf{0.057} & \textbf{8.28} & \textbf{19.96} \\
\bottomrule
\end{tabular}
}
\end{threeparttable}
\caption{Results of 12-12 multi-step prediction on four datasets evaluated using MAE and RMSE. The results presented in the table are obtained by averaging errors across all prediction steps. \textbf{Bold} denotes the best results, and \underline{underline} indicates the second-best results.}
\label{tbl:12-12}
\vspace{-2mm}
\end{table*}

\subsubsection{Fusion of Noise Priors}
According to Eq.~(\ref{eq:prior-target}) and Eq.~(\ref{eq:reference}), when the dynamics are close to the target values (i.e., \(\Delta x\) is small),  the noise prior \( \widetilde{\epsilon}\) can serve as a reference noise to \(\epsilon\).  Considering that the extracted prior dynamics represent regular patterns and may not always be accurate enough when facing irregular changes, we still need the denoising network to estimate the residual noise. So we define the final noise \(\widehat{\epsilon}\) as a weighted combination of the noise prior \( \widetilde{\epsilon}\) and the model-predicted noise \(\epsilon_{\theta}(x_n^{ta}, n| x_0^{co})\). This combination is controlled by an adjustable hyperparameter \(\lambda\) that balances the contributions of the noise prior and residual noise: 
\begin{equation}
\label{eq:fusion}
    \widehat{\epsilon}=\lambda\widetilde{\epsilon} + (1-\lambda)\epsilon_{\theta}(x_n^{ta}, n| x_0^{co}).
\end{equation}

By applying Eq.~(\ref{eq:fusion}), the noise prior is combined with the model's output, allowing the integration of data dynamics by manipulating the noise in the diffusion process.

\subsection{Training and Inference}
We provide the detailed algorithmic procedures for training and inference in Appendix~\ref{app: train_infer}. Different from the vanilla diffusion models, the optimization objective in the training phase has shifted from Eq.~(\ref{eq:loss1}) to:
\begin{equation}
\begin{aligned}
       \min_{\theta}\mathcal{L}(\theta)  = \min_{\theta} \mathbb{E}_{n, x_0, \epsilon} \left[ \left\| \epsilon - \widehat{\epsilon}\right\|_2^2 \right].
\end{aligned}
\end{equation}
\noindent In the sampling phase, the expression of \(\mu_{\theta}(x_n^{ta}, n| x_0^{co})\) in Eq~(\ref{eq:inference}) is reformulated as:
\begin{equation}
\begin{aligned}
\label{eq:denoising}
    &\mu_{\theta}(x_n^{ta}, n| x_0^{co})=\frac{1}{\sqrt{{\alpha}_n}} \left( x_n^{ta}-\frac{\beta_n}{\sqrt{1 - \bar{\alpha}_n}}\widehat{\epsilon} \right).
\end{aligned}
\end{equation}

\subsection{Flexibility of NPDiff}
NPDiff offers flexibility and universality by seamlessly integrating with state-of-the-art denoising network architectures.
To showcase its adaptability, we employ three representative models from the spatio-temporal domain as the denoising networks: CSDI ~\cite{tashiro2021csdi}, ConvLSTM ~\cite{shi2015convolutional}, and STID ~\cite{shao2022spatial}. These models were carefully selected to represent a diverse spectrum of deep learning paradigms. Specifically, CSDI leverages transformers as its denoising backbone, ConvLSTM incorporates Convolutional LSTM structures, and STID employs a multi-layer perceptron (MLP)-based architecture.   
Together, these models encompass the core building blocks of modern deep learning—namely CNNs, LSTMs, MLPs, and transformers—ensuring that NPDiff's applicability extends across various network modules.

\section{Evaluations} \label{section: experiment}
\subsection{Experimental Settings}
We provide a detailed description of the datasets, baselines, and experimental configurations in Appendix~\ref{app:A}.

\noindent\textbf{Datasets.}\label{section: data} We evaluate our method on four large-scale cellular network traffic datasets: MobileBJ, MobileNJ, MobileSH14 and MobileSH16, which are sourced from three major cities in China: Beijing, Nanjing, and Shanghai. 

\noindent\textbf{Baselines.} We select 13 state-of-the-art models as our baselines, which are divided into three categories:
\textit{\textbf{Classic models}} include HA and ARIMA, which can be applied to time series prediction without extensive training. \textit{\textbf{Time series prediction models}} include PatchTST~\cite{nie2022time}, iTransformer~\cite{liu2023itransformer}, and Time-LLM~\cite{jin2023time}, which are state-of-the-art models for multivariate time series forecasting. \textit{\textbf{Deep spatio-temporal prediction models}} consist of two groups: \textit{(i) urban spatio-temporal models}, including STResNet~\cite{zhang2017deep}, ATFM~\cite{liu2018attentive}, STNorm~\cite{deng2021st}, STGSP~\cite{zhao2022st}, TAU~\cite{tan2023temporal}, and PromptST~\cite{zhang2023promptst}; \textit{(ii) video prediction models}, including MAU~\cite{chang2021mau} and MIM~\cite{wang2019memory}, considering that the format of mobile traffic data is similar to video data.

\subsection{Overall Performance}
Our experiments include both multi-step and one-step prediction tasks. The multi-step prediction evaluates capabilities of long-term prediction, while the one-step prediction task assesses
the model’s ability to tackle abrupt changes. 

\begin{table*}[t!]
\centering
\setlength\tabcolsep{4mm}
\vspace{-2mm}


\vspace{-2mm}
\begin{threeparttable}
\resizebox{1.7\columnwidth}{!}{
\begin{tabular}{ccccccccc}
\toprule
\multirow{2}{*}{\textbf{Model}}
& \multicolumn{2}{c}{\textbf{MobileBJ}} & \multicolumn{2}{c}{\textbf{MobileNJ}} & \multicolumn{2}{c}{\textbf{MobileSH14}} & \multicolumn{2}{c}{\textbf{MobileSH16}} \\
\cmidrule(lr){2-3} \cmidrule(lr){4-5} \cmidrule(lr){6-7} \cmidrule(lr){8-9}
 & \textbf{MAE} & \textbf{RMSE} & \textbf{MAE} & \textbf{RMSE} & \textbf{MAE} & \textbf{RMSE} & \textbf{MAE} & \textbf{RMSE} \\
\midrule
CSDI & 0.251 & 0.824 & 0.202 & 0.612 & 0.043 & 0.068 & 11.20 & 45.98 \\
\textbf{+Periodic Prior} & 0.090 & 0.167 & 0.099 & 0.160 & 0.037 & 0.057 & 6.78 & 17.25 \\
\textbf{+Local Prior} & 0.059 & 0.094 &  0.097 &  0.196 & 0.034 &  0.056 & 5.36 & 9.74 \\
\midrule
ConvLSTM & 0.080 & 0.127 & 0.117 & 0.234 & 0.042 & 0.065 & 7.23 & 18.93 \\
\textbf{+Periodic Prior} & 0.085 & 0.157 & 0.098 & 0.157 & 0.036 & 0.054 & 8.21 & 19.34 \\
\textbf{+Local Prior} & 0.063 & 0.101 &  0.109 &  0.217 & 0.034 &  0.056 & 5.36 & 9.95  \\

\midrule
STID & 0.077 & 0.125 & 0.096 & 0.168 & 0.040 & 0.062 & 7.02 & 13.73 \\
\textbf{+Periodic Prior} & 0.089 & 0.159 & 0.098  & 0.166 & 0.036 & 0.055 & 7.24 & 17.30 \\
\textbf{+Local Prior} & 0.061 & 0.095 & 0.088 & 0.163  & 0.033 &  0.055 & 5.48 & 10.27  \\
\bottomrule
\end{tabular}}
\end{threeparttable}
\caption{Results of 12-1 one-step prediction on four datasets evaluated using MAE and RMSE.}
\label{tbl:12-1}
\end{table*}



\noindent\textbf{Multi-step Prediction.}
For the main experiment, we use 12 historical steps to predict the next 12 steps, which is one of the most typical setups in mobile traffic prediction. It is worth noting that different datasets have different temporal resolutions, so the time length corresponding to 12 steps differs. 
Table~\ref{tbl:12-12} shows the prediction results of NPDiff with CSDI (results for ConvLSTM and STID are shown in Appendix~Table~\ref{tbl:12-12_}).
Specifically, we have the following observations:

\begin{itemize}[leftmargin=*]
  
    \item \textit{\textbf{Video prediction model performs poorly on mobile traffic data.}} MAU and MIM, show relatively poor performance compared to other models, indicating that the dynamics of video data differ from those of mobile traffic data. Although video data and mobile traffic data share structural similarities, directly utilizing video prediction models delivers unsatisfying performance.
    \item \textit{\textbf{Our proposed noise prior framework achieves superior performance.}} Our proposed approach achieves the best performance compared to competitive baselines across almost all datasets. Compared with CSDI without noise prior, it has significant improvements in most cases, with an average reduction of \textbf{41.6\%} in MAE and \textbf{54.4\% }in RMSE. The remarkable performance improvement highlights the great potential of manipulating noise in diffusion models.
    \item \textit{\textbf{More accurate prior dynamics boost more performance gain.}} Specifically, on the MobileBJ and MobileNJ datasets, the diffusion models' average performance improves by 40.6\% and 47.2\% in terms of MAE when using the noise prior. Meanwhile, on the MobileSH14 and MobileSH16 datasets, although the improvements are slightly lower, there is still an average performance increase of 21.1\% and 25.5\% in MAE. The observations align with the cosine similarity results shown in Table~\ref{tbl:cos_similarity}, where the similarity scores for MobileBJ and MobileNJ datasets are higher.

\end{itemize}

\begin{figure}[t]
    \centering
    \includegraphics[width=\linewidth]{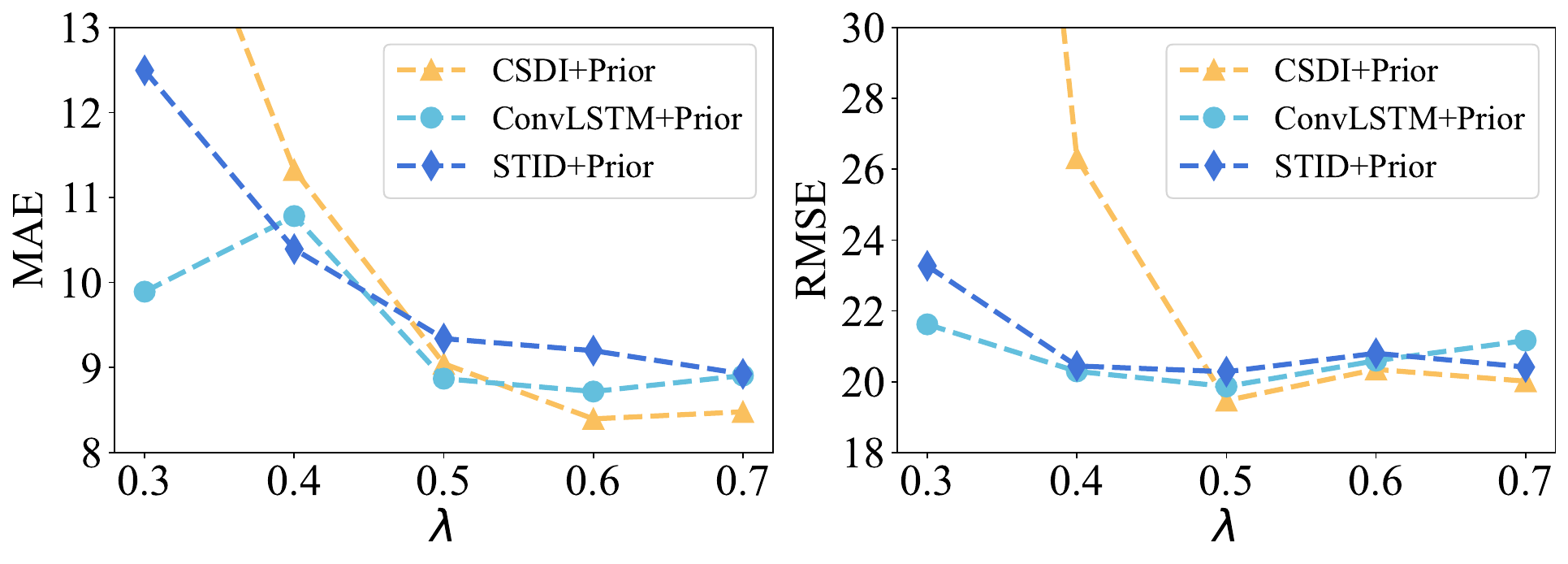}
    \caption{Ablation studies of $\lambda$  on the MobileSH16 dataset.}
    \label{fig:lambda SH}
\end{figure}

\begin{figure}[t]
    \centering
    \includegraphics[width=\linewidth]{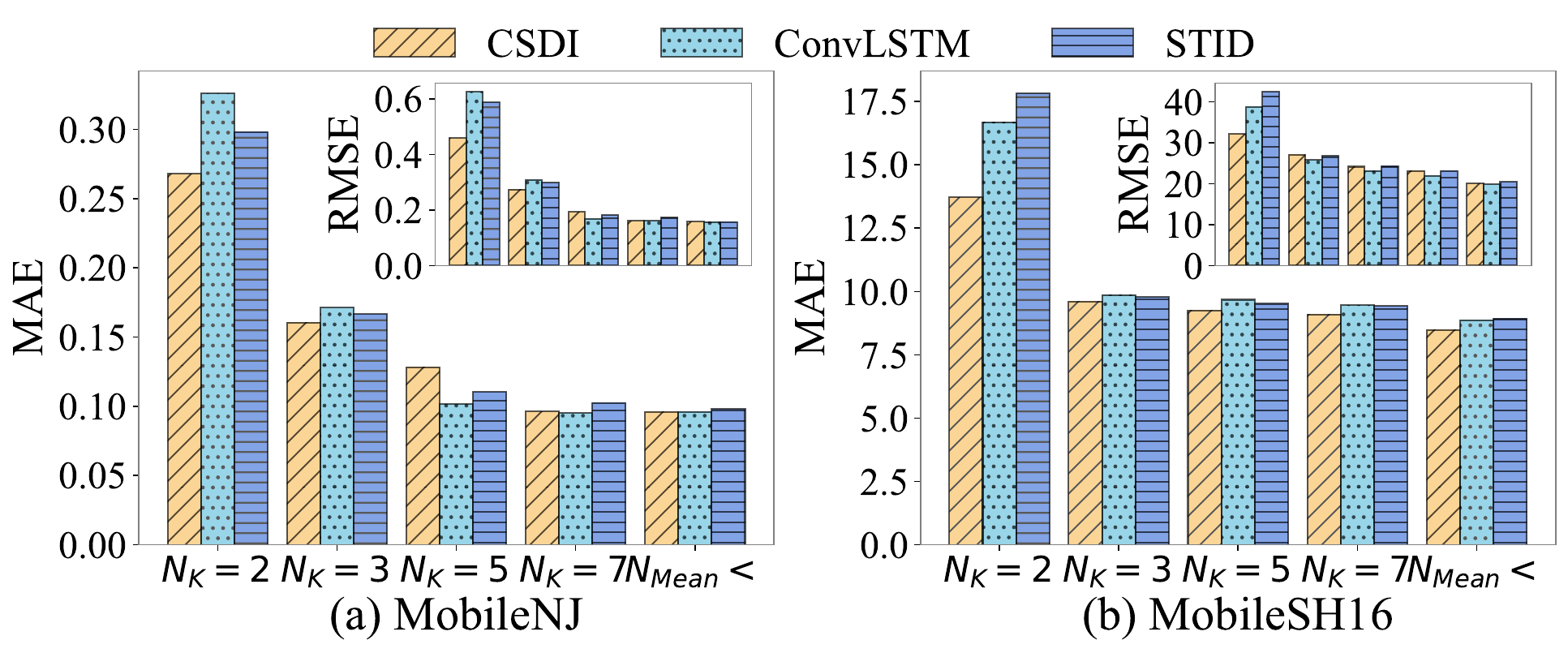}
    \vspace{-5mm}
    \caption{Ablation studies on the number of FFT frequency components in periodic dynamics.}
    \label{fig:comp}
    \vspace{-2mm}
\end{figure}

\noindent\textbf{One-step Prediction.}
For this task, we set the context length to 12 and limit the prediction horizon to 1. Specifically, we use periodic dynamics and local dynamics, separately, to investigate these contributions to the final performance. Table~\ref{tbl:12-1} illustrates the results of one-step prediction. When applying periodic dynamics, significant performance gains are observed in most datasets, with an average MAE reduction of 30.3\%. However, in a few datasets, it does not yield expected performance. In contrast, local dynamics proves effectiveness across all datasets and reduces the MAE by 35.1\% on average. 
It indicates that local dynamics are more suitable for one-step prediction, due to their ability to reflect short-term abrupt and sudden changes.
Also as shown in Table~\ref{tbl:cos_similarity}, the similarity between local dynamics and the real data is higher than that of periodic dynamics, indicating the importance of adopting suitable dynamics as noise prior.

\subsection{Ablation Studies}
Noise prior is the most essential design of our method. In this section, we explore the impact of two key hyperparameters associated with it: $N_k$ and $\lambda$.

\noindent\textbf{Noise Fusion Coefficient.}
Figure~\ref{fig:lambda SH} shows the impact of the coefficient $\lambda$ in Eq.~(\ref{eq:fusion}). We can observe that as $\lambda$ increases, the model’s performance gradually improves, demonstrating the effectiveness of noise prior. This also indicates that increasing the contribution of the noise prior allows for better utilization of prior dynamics. The model achieves optimal performance when $\lambda$ is around 0.5. However, as $\lambda$ continues to increase, a slight performance decline occurs. This can be attributed to the fact that the extracted prior dynamics only provide regular dynamics and cannot fully represent all variations in the real data. Over-reliance on the noise prior may result in suboptimal performance.

\noindent\textbf{Components of FFT.} Figure~\ref{fig:comp} illustrates the impact of the number of FFT frequency components in periodic dynamics on the model’s performance. The experimental results show that as the number of components increases, the performance of models improves accordingly. When the number of Fourier components reaches 5, allowing for sufficiently accurate capture of the data dynamics, the model’s performance metrics tend to stabilize.

\subsection{Training Efficiency and Robustness}

\begin{figure}[t]
    \centering
    \vspace{-3mm}
    \includegraphics[width=\linewidth]{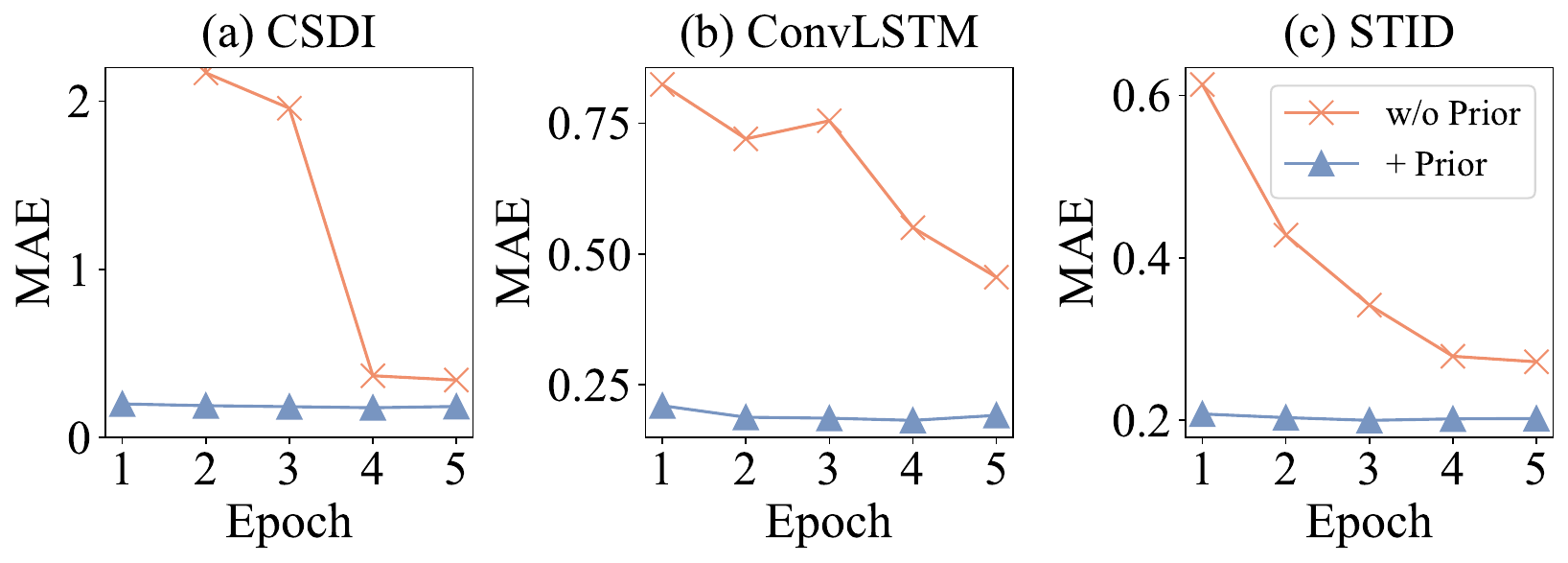}
    \vspace{-4mm}
    \caption{Comparison of training efficiency on the validation set.}
    \label{fig:val mae}
\end{figure}

\noindent\textbf{Training Efficiency.}
Figure~\ref{fig:val mae} illustrates the trend of performance on the validation set over the first five epochs. We can observe that the model with noise prior exhibits significantly faster convergence in the early stages of training. This rapid convergence can be attributed to the model’s ability to directly estimate part of the noise, enabling it to reach the optimal solution more quickly during training. This efficiency improvement is important for real-time applications.

\begin{figure}[t]
    \centering
    \vspace{-2mm}
    \includegraphics[width=0.65\linewidth]{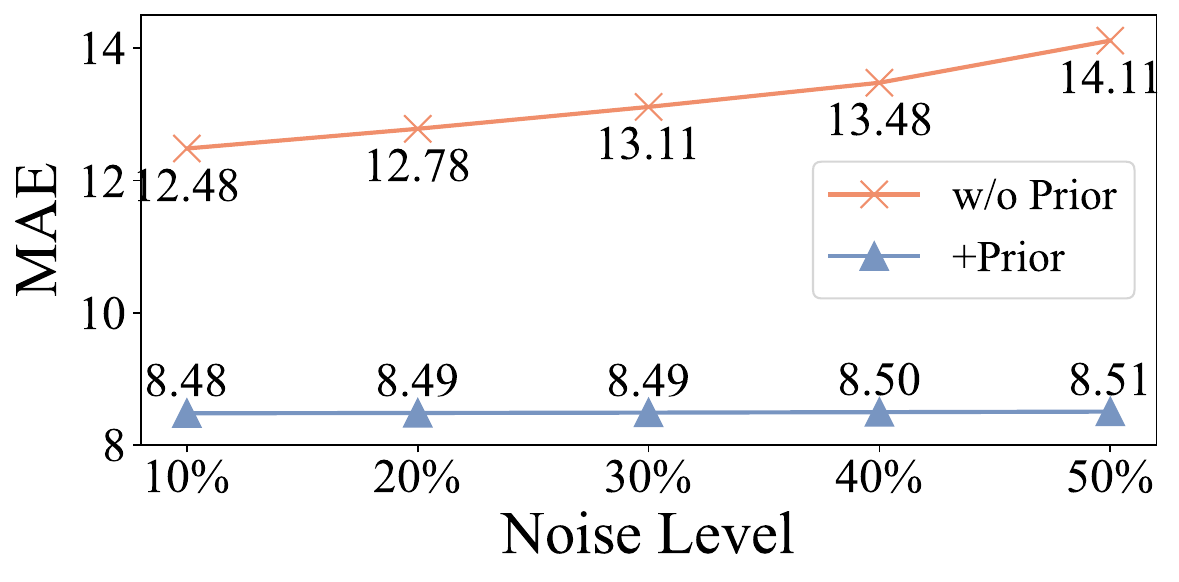}
    \vspace{-2mm}
    \caption{Results of noise perturbation on the MobileSH16 dataset.}
    \label{fig:Noise robutness}
\end{figure}

\noindent\textbf{Robustness Analysis.} We examine the robustness of NPDiff from two aspects: 

\begin{itemize}[leftmargin=*]
    \item \textit{\textbf{Different Prediction Tasks.}} We extend the evaluation task to a series of scenarios with different historical horizons and prediction steps, including 12-6, 24-24, 24-12, and 24-1. The detailed results can be found in Appendix~\cref{tbl:12-6,tbl:24-24,tbl:24-12,tbl:24-1}. The results show that our method improves model performance across all prediction tasks, demonstrating its robustness in different tasks.
    \item \textit{\textbf{Noise Perturbations.}} We evaluate the performance of our method under noise perturbations by introducing Gaussian noise with a mean of 0 into the mobile traffic data. The variance levels are set to range from 10\% to 50\% of the mean value. Figure~\ref{fig:Noise robutness} shows the performance of CSDI under different levels of noise perturbation. The results show that as the noise level gradually increases, the models incorporating noise prior exhibit higher stability, with only a minor performance decrease, proving its robustness.
\end{itemize}

\begin{figure}[t]
    \centering
    \includegraphics[width=\linewidth]{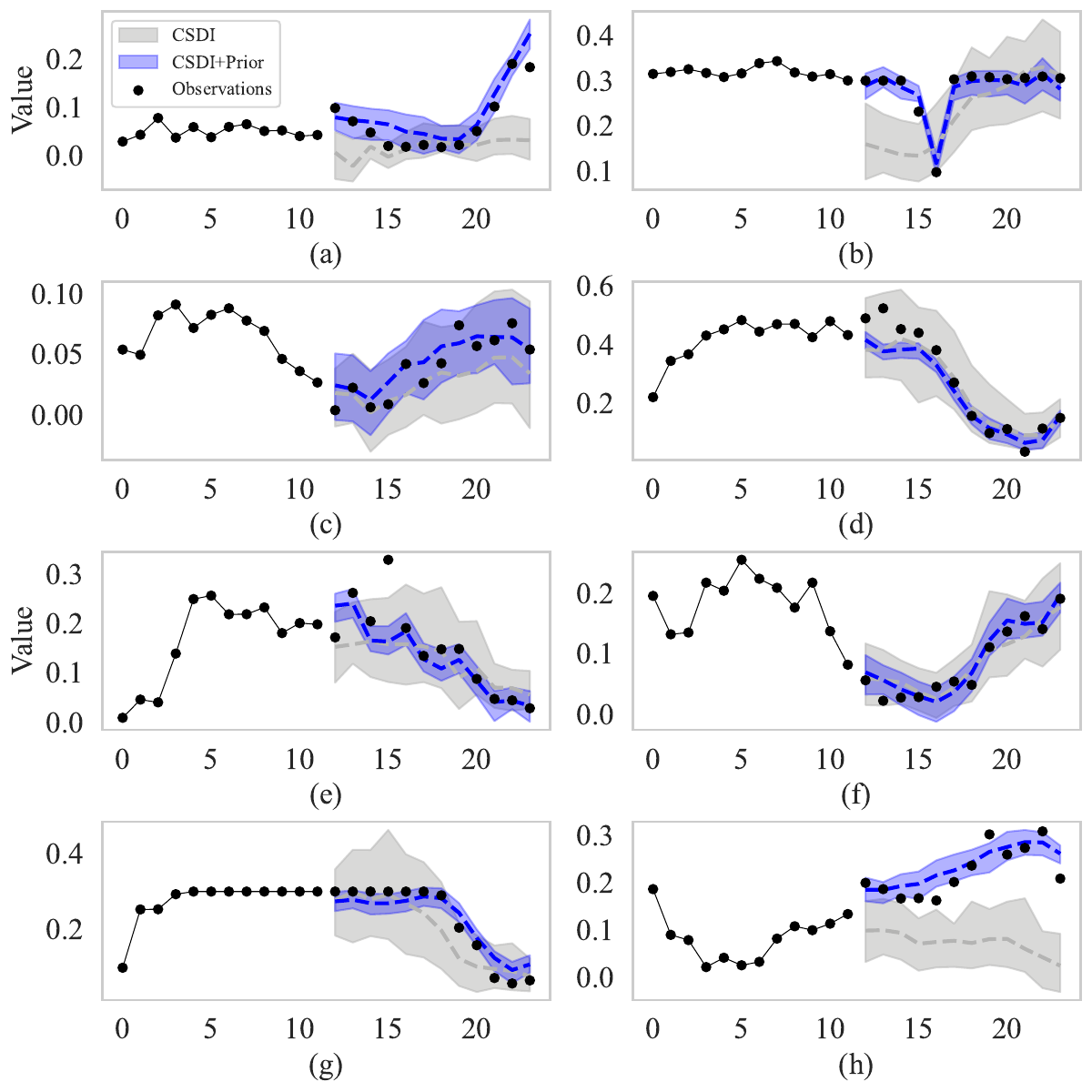}
    \caption{Visualizations of prediction uncertainties using CSDI and CSDI with noise prior on the MobileNJ dataset. The shaded areas represent prediction uncertainty, illustrating the 90\% confidence interval based on 50 independent runs. The dashed lines indicate the median of the prediction results for each model.}
    \label{fig:estimation}
\end{figure}

\subsection{Case Study}

To further investigate the impact of noise prior, we visualize the prediction results for the MobileNJ dataset. We use CSDI as an example, and Figure~\ref{fig:estimation} illustrates the results (the prediction results with ConvLSTM and STID can be found in Appendix~Figure~\ref{fig:estimation_stid_conv}). 

We observe that, compared to CSDI (indicated in gray), incorporating the noise prior (shown in blue) significantly reduces the uncertainty in predictions. This is especially important for probabilistic forecasting, where uncertainty reflects the reliability of the model’s predictions.
Additionally, as shown in Figure~\ref{fig:estimation}(a, b, and h), when dealing with uncommon circumstances, where sudden changes occur, our approach can well handle them due to the incorporation of prior dynamics.

\section{Conclusion}

In this paper, we propose NPDiff, a general noise prior framework for diffusion-based mobile traffic prediction, pioneering the manipulation of noise in diffusion models for temporal data prediction. NPDiff offers significant advancements over existing solutions in several key areas, including enhanced prediction accuracy, improved training efficiency, increased robustness, and reduced uncertainty. These improvements provide a more efficient and reliable approach to real-world mobile traffic management, demonstrating the framework's potential to enhance both performance and practical applicability in this critical domain.

Our work is highly extensible, offering multiple avenues for future exploration. On one hand, further research could focus on refining the fusion of the noise prior and residual noise to better handle complex data scenarios. On the other hand, designing more advanced noise priors is another promising direction.  This could involve developing priors that adapt to the unique dynamics of various spatio-temporal scenarios or creating a more general prior that can generalize effectively across diverse scenarios. 

\clearpage
\section*{Acknowledgements}
This work is supported in part by the National Natural Science Foundation of China under Grants 62476152 and U24B20180.
\bibliographystyle{named}
\bibliography{ijcai25}
\clearpage

\section*{APPENDIX}

\appendix
\begin{table*}[t!]
\centering
\vspace{-1mm}

\begin{threeparttable}
\resizebox{1.75\columnwidth}{!}{
\begin{tabular}{cccccccc}
\toprule
\textbf{Dataset} & \textbf{City}  & \textbf{Temporal Period} & \textbf{Spatial partition} & \textbf{Interval} & \textbf{Mean} & \textbf{Std} \\
\hline
MobileBJ & Beijing  &  2021/10/25 - 2021/11/21 & $28 \times 24$ & One hour & 0.367 & 0.411 \\
MobileNJ & Nanjing  &  2021/02/02 - 2021/02/22 & $20 \times 28$ & One hour & 0.842 & 1.299 \\
MobileSH14 & Shanghai  &  2014/08/01 - 2014/08/21 & $32 \times 28$ & One hour & 0.175 & 0.212 \\
MobileSH16 & Shanghai  &  2016/04/25 - 2016/05/01 & $20 \times 20$ & 15min & 31.935 & 137.926 \\

\bottomrule
\end{tabular}}
\end{threeparttable}
\caption{The basic information of four mobile traffic datasets.}
\label{tbl:append_data_grid}
\end{table*}

\begin{table*}[t!]
\centering

\begin{threeparttable}
\resizebox{1.5\columnwidth}{!}{
\begin{tabular}{ccccccccc}
\toprule
\multirow{2}{*}{\textbf{Model} }

& \multicolumn{2}{c}{\textbf{MobileBJ}} & \multicolumn{2}{c}{\textbf{MobileNJ}} & \multicolumn{2}{c}{\textbf{MobileSH14}} & \multicolumn{2}{c}{\textbf{MobileSH16}}  \\
\cmidrule(lr){2-3} \cmidrule(lr){4-5} \cmidrule(lr){6-7} \cmidrule(lr){8-9} 
 &\textbf{MAE} & \textbf{RMSE} & \textbf{MAE} & \textbf{RMSE} & \textbf{MAE} & \textbf{RMSE} & \textbf{MAE} & \textbf{RMSE} \\
\midrule
ConvLSTM & 0.123 & 0.205 & 0.216 & 0.489 & 0.050 & 0.081 & 15.56 & 45.83 \\
\textbf{ConvLSTM+Prior} & 0.096 & 0.176 & 0.096 & 0.157 & 0.037 & 0.057 & 8.30 & 19.78 \\
\midrule
STID & 0.116 & 0.183 & 0.168 & 0.319 & 0.046 & 0.073 & 9.91 & 27.80 \\
\textbf{STID+Prior} & 0.098 & 0.176 & 0.098 & 0.156 & 0.037 & 0.057 & 8.93 & 20.41 \\

\bottomrule
\end{tabular}
}
\end{threeparttable}
\caption{Results of 12-12 multi-step prediction on four datasets evaluated using MAE and RMSE for ConvLSTM and STID models. The results presented in the table are obtained by averaging prediction errors across all prediction steps.}
\label{tbl:12-12_}
\end{table*}

\begin{table*}[t!]
\centering


\begin{threeparttable}
\resizebox{1.5\columnwidth}{!}{
\begin{tabular}{ccccccccc}
\toprule
\multirow{2}{*}{\textbf{Model}}
& \multicolumn{2}{c}{\textbf{MobileBJ}} & \multicolumn{2}{c}{\textbf{MobileNJ}} & \multicolumn{2}{c}{\textbf{MobileSH14}} & \multicolumn{2}{c}{\textbf{MobileSH16}} \\
\cmidrule(lr){2-3} \cmidrule(lr){4-5} \cmidrule(lr){6-7} \cmidrule(lr){8-9}
 & \textbf{MAE} & \textbf{RMSE} & \textbf{MAE} & \textbf{RMSE} & \textbf{MAE} & \textbf{RMSE} & \textbf{MAE} & \textbf{RMSE} \\
\midrule
CSDI & 0.153 & 0.483 & 0.202 & 0.638 & 0.068 & 0.112 & 9.74 & 29.61 \\
    
\textbf{CSDI+Prior} & 0.093 & 0.173 & 0.092 & 0.154 & 0.037 & 0.057 & 8.73 & 20.97 \\
\midrule
ConvLSTM & 0.139 & 0.227 & 0.172 & 0.364 & 0.044 & 0.069 & 10.90 & 32.29 \\
\textbf{ConvLSTM+Prior} & 0.093 & 0.168 & 0.107 & 0.164 & 0.037 & 0.057 & 8.32 & 20.39 \\
\midrule
STID & 0.102 & 0.168 & 0.132 & 0.257 & 0.045 & 0.071 & 8.94 & 21.71 \\
\textbf{STID+Prior} & 0.096 & 0.171 & 0.109 & 0.168 & 0.037 & 0.057 & 8.84 & 20.45 \\
\bottomrule
\end{tabular}}
\end{threeparttable}
\caption{Results of 12-6 multi-step prediction on four datasets evaluated using MAE and RMSE. The results presented in the table are obtained by averaging prediction errors across all prediction steps.}
\label{tbl:12-6}
\vspace{-2mm}
\end{table*}

\section{Implementation Details}\label{app:A}
\subsection{\textbf{Datasets}} \label{app:datasets}
Table~\ref{tbl:append_data_grid} provides basic information about the datasets. For our experiments, we divide each dataset into three parts with a ratio of 6:2:2. The first 60\% is used as the training set, the subsequent 20\% as the validation set, and the remaining 20\% as the test set. All datasets are standardized to a standard normal distribution. It is necessary to specify that the periodic dynamics utilized in our study are solely obtained through the analysis of the training set.

\subsection{\textbf{Experimental Configuration}} \label{app:settings}
For the model training, we set the maximum number of epochs to 100, accompanied by an early stopping strategy. For the validation, we sample 3 times and calculate the average result for early stopping. In the testing stage, we sample 50 times and report the average result. For baseline models, we set the maximum number of epochs to 200 for training. We provide the details of datasets, evaluation metrics, and experimental settings in the Appendix. For denoising networks, we set CSDI's residual layers to 4, with 64 residual channels and 8 attention heads. For ConvLSTM, we configure 3 LSTM layers with a hidden size of 64. For STID, we set encoder to 4 layers with an embedding dimension of 64. During training, we employ a quadratic noise schedule in our diffusion model, starting with a noise level of $\beta_1 = 0.0001$, which increases to a maximum of $\beta_N = 0.5$ over $N = 50$ steps. We use a batch size of 8 and an initial learning rate of 1e-3, which is reduced to 4e-4 after 40 epochs. We apply the Adam optimizer with a weight decay of 1e-6. For our method's hyperparameters, we set the number of periodic dynamic components $N_K$ to 5, or $N_m$, representing the number of components with amplitudes above the average. We select the weighted coefficient $\lambda$ from the range \{0.3, 0.4, 0.5, 0.6, 0.7\}. In the experiments, we report the best performance selected from all combinations of these two parameter settings.

\subsection{\textbf{Denoising Networks}} A brief introduction of the three denoising networks used in our experiments is provided below:

\begin{itemize}[leftmargin=*]
    \item CSDI~\cite{tashiro2021csdi}: CSDI is one of the most classic diffusion-based models in the spatio-temporal domain. Its denoising network is designed based on a transformer architecture and has demonstrated outstanding performance in time series data imputation tasks.
    \item ConvLSTM~\cite{shi2015convolutional}: ConvLSTM is a seminal model in spatio-temporal forecasting, utilizing Convolutional LSTM units to effectively capture spatio-temporal dynamics. It has demonstrated great performance in applications such as weather forecasting and traffic flow prediction. In our work, we utilize ConvLSTM as one of the denoising networks.
    \item STID~\cite{shao2022spatial}: STID is an efficient and straightforward spatio-temporal forecasting model that employs MLP to capture dependencies in multivariate time series, thereby avoiding the complexities inherent in deeper models. It effectively captures spatio-temporal features, demonstrating superior performance across various tasks. In our experiments, STID is utilized as a denoising network to develop a new diffusion-based model and to evaluate the performance of our approach.
\end{itemize}

\subsection{Training and Inference Algorithm}\label{app: train_infer}
The training process of NPDiff is shown in Algorithm~\ref{al: train}, and the sampling process is shown in Algorithm~\ref{al: infer}.

\begin{algorithm}[!t]
\caption{Training of NPDiff}
\label{al: train}
\begin{algorithmic}[1]
    \STATE \textbf{Input:} Distribution of training data $q(x_0)$, noise schedule ${\beta_n}$, data dynamics $D$
    \STATE \textbf{Output:} Trained denoising function $\epsilon_{\theta}$
    \REPEAT 
        \STATE Separate the values of $x_0$ into context data $x_0^{co}$  and target data $x_0^{ta}$
        \STATE Initialize $n \sim 
        \text{Uniform}(1, \dots, N)$ and $\epsilon \sim \mathcal{N}(0, I)$
        \STATE Calculate noisy targets $x_n^{ta} = \sqrt{\bar{\alpha}_n} x_0^{ta} + \sqrt{1 - \bar{\alpha}_n} \epsilon$
        \STATE Calculate noise prior $\widetilde{\epsilon}=\frac{x_n^{ta}-\sqrt{\bar{\alpha}_n}D}{\sqrt{1-\bar{\alpha}_n}}$
        \STATE Take a gradient step on:
        \[
        \nabla_\theta \left[ \left\| \epsilon - (1-\lambda)\epsilon_{\theta}(x_n^{ta}, n| x_0^{co})-\lambda \widetilde{\epsilon}\right\|_2^2 \right]
        \]
    \UNTIL{converged}
\end{algorithmic}
\end{algorithm}

\begin{algorithm}[!t]
\caption{Sampling of NPDiff}
\label{al: infer}
\begin{algorithmic}[1]
    \STATE \textbf{Input:} Context data $x_0^{co}$, data dynamics $D$, trained denoising function $\epsilon_{\theta}$
    \STATE \textbf{Output:} Target data $x_0^{ta}$
    \STATE Sample $x_N^{ta}$ from $\epsilon \sim \mathcal{N}(0,I)$
    \FOR{$n = N$ to $1$} 
        \STATE Calculate noise prior $\widetilde{\epsilon}=\frac{x_n^{ta}-\sqrt{\bar{\alpha}_n}D}{\sqrt{1-\bar{\alpha}_n}}$
        \STATE Predict noise $\epsilon_\theta(x_{n}^{ta},n|x_0^{co})$
        \STATE Reverse diffusion to get $x_{n-1}^{ta}$ using Eq. (\ref{eq:denoising})
        
    \ENDFOR
    \STATE \textbf{Return:} $x_0^{ta}$
\end{algorithmic}
\end{algorithm}

\section{Baselines} \label{app: baselines}

\begin{itemize}[leftmargin=*]

    \item \textbf{HA}: The History Average approach relies on the mean of data from previous time intervals to forecast target values.
    \item \textbf{ARIMA}: The ARIMA model is a frequently applied statistical approach for time series forecasting. This method effectively predicts time series data obtained at consistent time intervals.

    \item \textbf{PatchTST}~\cite{nie2022time}: It introduces patching and self-supervised learning for multivariate time series forecasting. The time series is divided into segments to capture long-term dependencies. Different channels are processed independently using a shared network. 
    \item \textbf{iTransformer}~\cite{liu2023itransformer}: This advanced model for multivariate time series prediction leverages attention and feed-forward processes on an inverted dimension. It focuses on capturing the relationships between different variables.
    
    \item \textbf{Time-LLM}~\cite{jin2023time}: TIME-LLM represents a leading-edge approach for applying large language models to time series forecasting. It employs a reprogramming framework that adapts LLMs for general time series predictions while maintaining the original structure of the language models.

    \item \textbf{STResNet}~\cite{zhang2017deep}: It makes use of residual neural networks to model dynamics in the data. These networks effectively capture temporal closeness, periodic behaviors, and long-term trends. 
    \item \textbf{ATFM}~\cite{liu2018attentive}: The Attentive Crowd Flow Machine model forecasts crowd movement by using an attention mechanism. This mechanism adaptively combines both sequential and periodic dynamics to improve prediction accuracy.
    \item \textbf{STNorm}~\cite{deng2021st}: It introduces two distinct normalization modules: spatial normalization and temporal normalization. These modules work independently, with spatial normalization managing high-frequency components and temporal normalization addressing local variations.
    \item \textbf{STGSP}~\cite{zhao2022st}: This model underscores the importance of both global and positional information within the temporal dimension when performing spatio-temporal predictions. By employing a semantic flow encoder, it captures temporal positional signals, while an attention mechanism is used to handle multi-scale temporal dependencies.
    
    \item \textbf{TAU}~\cite{tan2023temporal}: The Temporal Attention Unit breaks down temporal attention into intra-frame and inter-frame components. It introduces differential divergence regularization to effectively manage variations between frames.
    \item \textbf{PromptST}~\cite{zhang2023promptst}: It is a state-of-the-art model that combines pre-training and prompt-tuning, specifically tailored for spatio-temporal prediction.
    
    \item \textbf{MIM}~\cite{wang2019memory}: This model utilizes differential information between adjacent recurrent states to address non-stationary characteristics. By stacking multiple MIM blocks, it enables the modeling of higher-order non-stationarity.
    \item \textbf{MAU}~\cite{chang2021mau}: The Motion-aware Unit improves the detection of motion correlations between frames by extending the temporal coverage of the prediction units. It employs both an attention mechanism and a fusion module to enhance video forecasting.

\end{itemize}

\begin{table*}[t!]
\centering


\begin{threeparttable}
\resizebox{1.5\columnwidth}{!}{
\begin{tabular}{ccccccccc}
\toprule
\multirow{2}{*}{\textbf{Model}}
& \multicolumn{2}{c}{\textbf{MobileBJ}} & \multicolumn{2}{c}{\textbf{MobileNJ}} & \multicolumn{2}{c}{\textbf{MobileSH14}} & \multicolumn{2}{c}{\textbf{MobileSH16}} \\
\cmidrule(lr){2-3} \cmidrule(lr){4-5} \cmidrule(lr){6-7} \cmidrule(lr){8-9}
 & \textbf{MAE} & \textbf{RMSE} & \textbf{MAE} & \textbf{RMSE} & \textbf{MAE} & \textbf{RMSE} & \textbf{MAE} & \textbf{RMSE} \\
\midrule
CSDI & 0.184 & 0.522 & 0.130 & 0.262 & 0.047 & 0.076 & 9.95 & 29.77 \\
\textbf{CSDI+Prior} & 0.089 & 0.168 & 0.097 & 0.168 & 0.037 & 0.057 & 8.98 & 21.02 \\
\midrule
ConvLSTM & 0.084 & 0.144 & 0.159 & 0.339 & 0.049 & 0.080 & 19.52 & 53.24 \\
\textbf{ConvLSTM+Prior} & 0.090 & 0.165 & 0.098 & 0.158 & 0.037 & 0.057 & 10.53 & 23.36 \\
\midrule
STID & 0.107 & 0.168 & 0.148 & 0.307 & 0.054 & 0.084 & 13.96 & 37.44 \\
\textbf{STID+Prior} & 0.089 & 0.166 & 0.102 & 0.163 & 0.037 & 0.057 & 11.56 & 26.76 \\
\bottomrule
\end{tabular}}
\end{threeparttable}
\caption{Results of 24-24 multi-step prediction on four datasets evaluated using MAE and RMSE. The results presented in the table are obtained by averaging prediction errors across all prediction steps.}
\label{tbl:24-24}
\end{table*}

\begin{table*}[t!]
\centering


\begin{threeparttable}
\resizebox{1.5\columnwidth}{!}{
\begin{tabular}{ccccccccc}
\toprule
\multirow{2}{*}{\textbf{Model}}
& \multicolumn{2}{c}{\textbf{MobileBJ}} & \multicolumn{2}{c}{\textbf{MobileNJ}} & \multicolumn{2}{c}{\textbf{MobileSH14}} & \multicolumn{2}{c}{\textbf{MobileSH16}} \\
\cmidrule(lr){2-3} \cmidrule(lr){4-5} \cmidrule(lr){6-7} \cmidrule(lr){8-9}
 & \textbf{MAE} & \textbf{RMSE} & \textbf{MAE} & \textbf{RMSE} & \textbf{MAE} & \textbf{RMSE} & \textbf{MAE} & \textbf{RMSE} \\ 


\midrule
CSDI & 0.156 & 0.364 & 0.139 & 0.301 & 0.049 & 0.078 & 14.62 & 66.96 \\
\textbf{CSDI+Prior} & 0.092 & 0.172 & 0.095 & 0.157 & 0.037 & 0.057 & 9.38 & 21.65 \\
\midrule

ConvLSTM & 0.105 & 0.172 & 0.381 & 0.803 & 0.051 & 0.082 & 13.72 & 39.45 \\
\textbf{ConvLSTM+Prior} & 0.092 & 0.169 & 0.097 & 0.161 & 0.037 & 0.057 & 10.37 & 24.07 \\
\midrule
STID & 0.084 & 0.131 & 0.133 & 0.268 & 0.050 & 0.079 & 10.70 & 30.99 \\
\textbf{STID+Prior} & 0.091 & 0.168 & 0.108 & 0.161 & 0.037 & 0.057 & 10.28 & 23.99 \\

\bottomrule
\end{tabular}}
\end{threeparttable}
\caption{Results of 24-12 multi-step prediction on four datasets evaluated using MAE and RMSE. The results presented in the table are obtained by averaging prediction errors across all prediction steps.}
\label{tbl:24-12}
\end{table*}

\begin{table*}[t!]
\centering


\begin{threeparttable}
\resizebox{1.5\columnwidth}{!}{
\begin{tabular}{ccccccccc}
\toprule
\multirow{2}{*}{\textbf{Model}}
& \multicolumn{2}{c}{\textbf{MobileBJ}} & \multicolumn{2}{c}{\textbf{MobileNJ}} & \multicolumn{2}{c}{\textbf{MobileSH14}} & \multicolumn{2}{c}{\textbf{MobileSH16}} \\
\cmidrule(lr){2-3} \cmidrule(lr){4-5} \cmidrule(lr){6-7} \cmidrule(lr){8-9}
 & \textbf{MAE} & \textbf{RMSE} & \textbf{MAE} & \textbf{RMSE} & \textbf{MAE} & \textbf{RMSE} & \textbf{MAE} & \textbf{RMSE} \\

\midrule
CSDI & 0.223 & 0.809 & 0.177 & 0.662 & 0.066 & 0.160 & 8.49 & 28.99 \\
\textbf{+Periodic Prior} & 0.087 & 0.157 & 0.106 & 0.162 & 0.037 & 0.057 & 8.12 & 19.92 \\
\textbf{+Local Prior} & 0.058 & 0.092 & 0.098 &  0.178 & 0.033 &  0.053 & 5.27 & 9.65 \\
\midrule
ConvLSTM & 0.085 & 0.139 & 0.113 & 0.213 & 0.040 & 0.063 & 7.22 & 18.26 \\
\textbf{+Periodic Prior} & 0.083 & 0.155 & 0.097 & 0.159 & 0.036 & 0.055 & 8.09  & 20.16  \\
\textbf{+Local Prior} & 0.064 & 0.101 &  0.117 &  0.220 & 0.034 & 0.055 & 5.26 & 9.81 \\
\midrule
STID & 0.131 & 0.171 & 0.105 & 0.180 & 0.039 & 0.058 & 6.52 & 13.73 \\
\textbf{+Periodic Prior} & 0.087 & 0.153 & 0.095 & 0.161 & 0.037 & 0.055 & 7.88 & 18.06 \\
\textbf{+Local Prior} & 0.051 & 0.079 & 0.094 &  0.174 & 0.031 &  0.050 & 5.47 & 10.31 \\
\bottomrule
\end{tabular}}
\end{threeparttable}
\caption{Results of 24-1 one-step prediction on four datasets evaluated using MAE and RMSE.}
\label{tbl:24-1}
\end{table*}

\section{Additional Results}

\begin{figure*}[t]
    \centering
    \vspace{-3mm}
    \includegraphics[width=\linewidth]{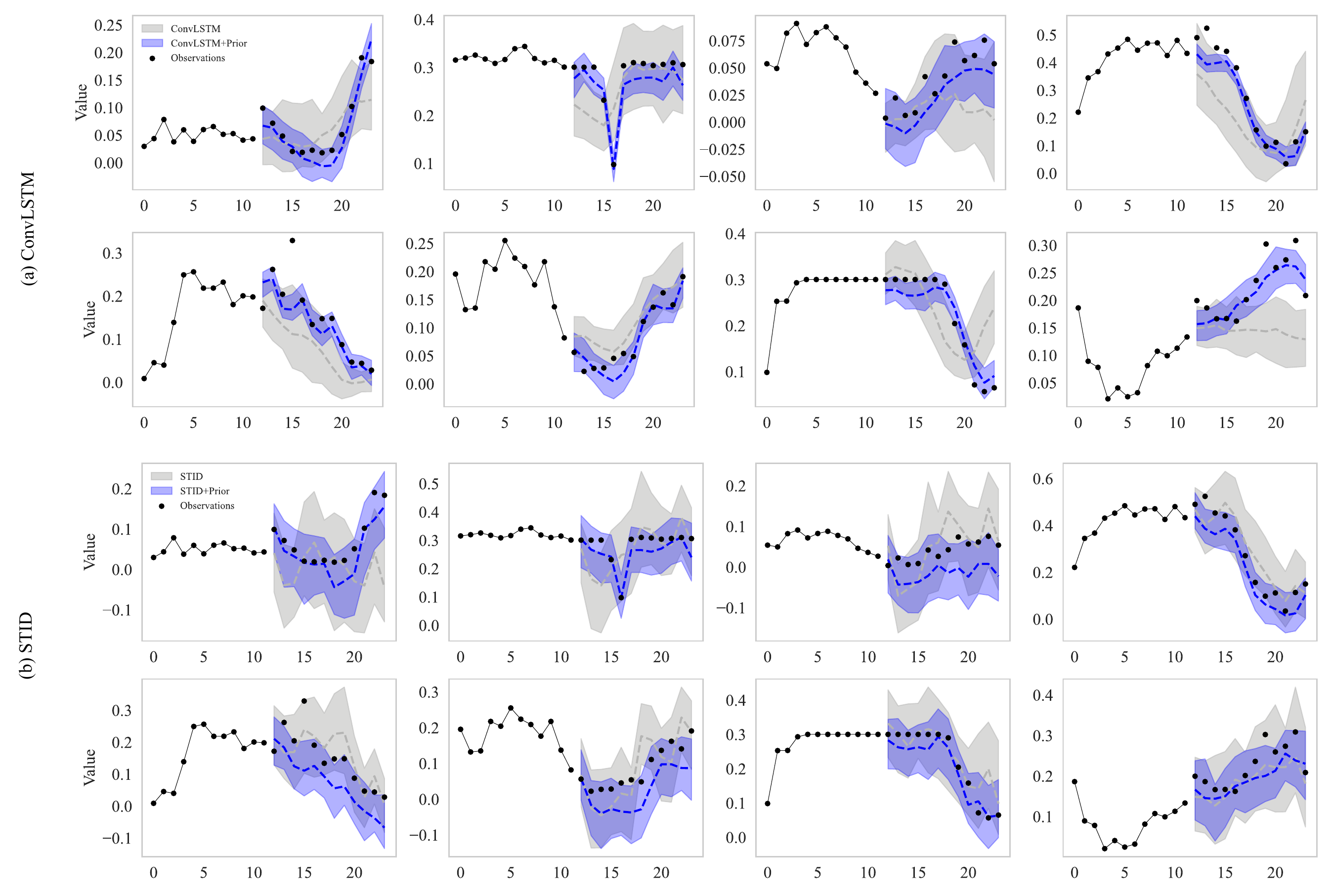}
    \vspace{-3mm}
    \caption{Prediction visualization comparing the noise prior-enhanced models and two baseline models (ConvLSTM and STID) on the MobileNJ dataset. The shaded areas represent prediction uncertainty, illustrating the 90\% confidence interval based on 50 independent runs. The dashed lines indicate the median of the predictions for each model.}
    \label{fig:estimation_stid_conv}
    \vspace{-3mm}
\end{figure*}

Table~\ref{tbl:12-12_} reports the results of two additional diffusion models for the 12-12 multi-step prediction task, both showing notable performance improvements across four datasets. \cref{tbl:12-6,tbl:24-24,tbl:24-12,tbl:24-1} detail the performance of our method across different prediction tasks, demonstrating strong outcomes in both multi-step and one-step predictions. Figure~\ref{fig:estimation_stid_conv} visualizes the prediction results with ConvLSTM and STID on MobileNJ dataset.





\begin{figure*}[t!]
    \centering
    \includegraphics[width=\linewidth]{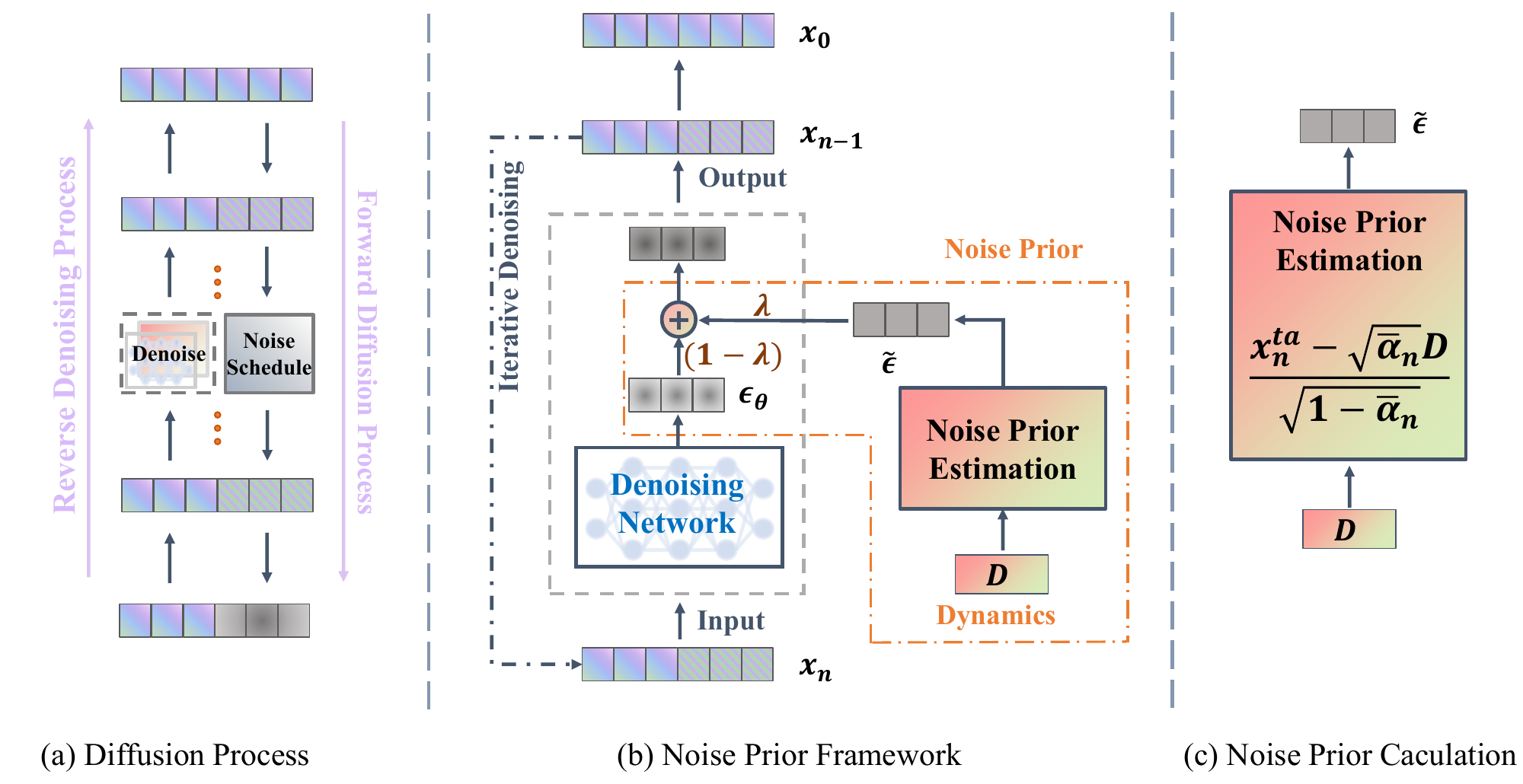}
    \caption{Detailed explanation of noise prior in the diffusion process.}
    \label{fig:revision_sup}
\end{figure*}

\end{document}